\pdfoutput=1

\documentclass[11pt]{article}

\usepackage[final]{acl}

\usepackage{amsfonts}
\usepackage{times}
\usepackage{latexsym}
\usepackage{tabularx} 
\usepackage{amsmath}
\usepackage{graphicx}
\usepackage{multirow}

\usepackage{pdflscape}
 
\usepackage[T1]{fontenc}

\usepackage[utf8]{inputenc}

\usepackage{microtype}

\usepackage{inconsolata}

\usepackage{makecell}

\usepackage{graphicx}

\title{Evolutionary Contrastive Distillation for Language Model Alignment}

\author{
 \textbf{Julian Katz-Samuels}, \,
 \textbf{Zheng Li}, \,
 \textbf{Hyokun Yun}, \\
 \textbf{Priyanka Nigam}, \,
 \textbf{Yi Xu}, \, 
 \textbf{Vaclav Petricek}, \\
 \textbf{Bing Yin}, \, 
 \textbf{Trishul Chilimbi} \\
 Amazon \\
  \small{
   \textbf{Correspondence:} \{jkatzsam,amzzhe,yunhyoku\}@amazon.com
 }}

\begin{document}
\maketitle
\begin{abstract}
The ability of large language models (LLMs) to execute complex instructions is essential for their real-world applications. However, several recent studies indicate that LLMs struggle with challenging instructions \citep{zhou2023instruction, qin2024infobench, jiang2023followbench}. In this paper, we propose Evolutionary Contrastive Distillation (ECD), a novel method for generating high-quality synthetic preference data designed to enhance
the complex instruction-following capability of language models. ECD generates data that specifically illustrates the difference between a response that successfully follows a set of complex instructions and a response that is high-quality, but nevertheless makes some subtle mistakes. This is done by prompting LLMs to progressively evolve simple instructions into more complex instructions.
When an instruction is made more complex, the original successful response mostly meets the new requirements but misses one or two, thus becoming a “hard negative” example for the new instruction.
By pairing a good response with such a hard negative response, and employing contrastive learning algorithms
such as DPO \citep{rafailov2023direct}, we improve language models' ability to follow complex instructions. Empirically, we observe that our method yields a 7B model that exceeds the complex instruction-following performance of current state-of-the-art (SOTA) 7B models and is competitive even with open-source 70B models.
\end{abstract}

\section{Introduction}

\begin{figure*}[htbp!]
  \centering
  \includegraphics[width=\textwidth]{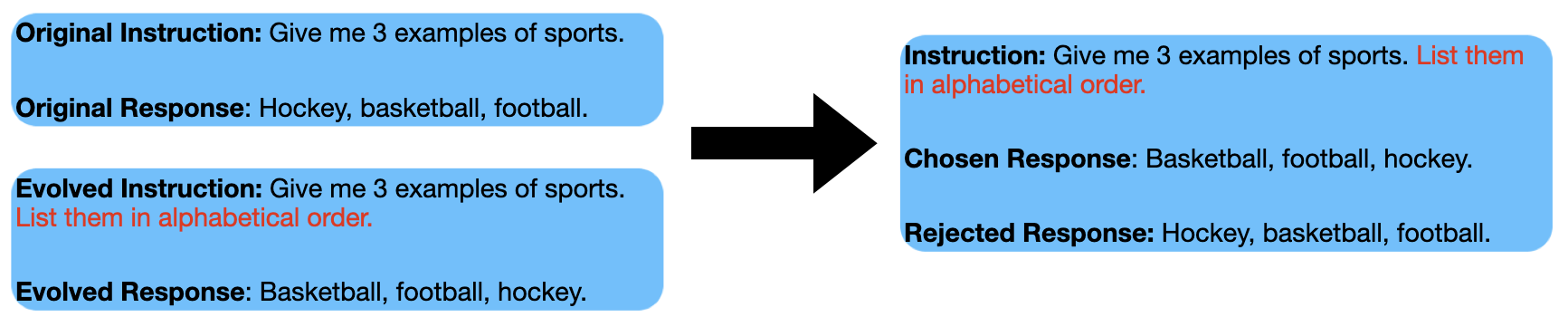}
  \caption{Example of ECD preference pair construction. The original instruction is to provide three examples of sports while the evolved instruction adds the additional requirement to sort the examples in alphabetical order. Since the original response violates this requirement, it can be used as a rejected response in the preference pair. On the other hand, the evolved response satisfies all the requirements and can be used as the chosen response.}
  \label{fig:rlced_data}
\end{figure*}

Large language models (LLMs) have demonstrated impressive capabilities in a wide range of tasks ranging from creative writing to code generation. In light of these achievements, there has been a surge of interest in building complex data processing systems with LLM-based components \citep{langchain, llamaindex}. Complex instruction-following capability, which is the capability of LLMs to generate outputs consistent with multiple interdependent specifications in the prompt, is critically important for such systems to operate reliably. Consequently, multiple benchmarks have been proposed to capture various aspects of complex instruction-following \citep{zhou2023instruction, qin2024infobench, jiang2023followbench}. 
Unfortunately, these studies find that there still is a significant gap between proprietary LLMs and open source models on these benchmarks.

This raises the question: how shall we effectively distill the complex instruction-following ability of stronger LLMs into smaller language models? After the seminal work by \citet{wang2023selfinstruct}, the usage of proprietary LLMs for the generation of alignment data has been actively studied. As a typical alignment pipeline consists of Supervised Fine-Tuning (SFT) methods and Preference Fine-Tuning (PFT\footnote{We denote this stage PFT instead of RLHF (Reinforcement Learning from Human Feedback), because many data collection methods for preference data do not necessarily involve human feedback \citep{lee2023rlaif, cui2023ultrafeedback, yang2023rlcd}.}) stages \citep{ouyang2022training}, these data generation methods are also largely categorized into SFT data generation methods (e.g., UltraChat \citep{ding2023enhancing}) and PFT data generation methods (e.g., UltraFeedback \citep{cui2023ultrafeedback}).

Among a wide variety of SFT data generatin methods,
Evol-Instruct \citep{xu2023wizardlm} and Conifer \citep{sun2024conifer} are specifically designed to improve complex instruction-following. They share the common goal: generate highly complex instructions. And they also share the same evolutionary strategy: they first prompt proprietary LLMs to evolve simple seed instructions from ShareGPT to have progressively more complex requirements. Then, proprietary LLMs are again prompted to generate demonstrations on these complex instructions.

On the other hand, while a wide variety of PFT data generation methods have been proposed in the literature \citep{lee2023rlaif, cui2023ultrafeedback, yang2023rlcd} there has been little attention on leveraging PFT stage for complex instruction-following. 
We claim, however, that PFT stage has distinctive advantages in teaching complex instruction-following.
First, SFT training places an equal weight on every token in the response, and this can be less than ideal for instruction-following. When a tricky requirement such as "make sure every sentence rhymes with each other" is imposed in the instruction, certain tokens play a more important role in meeting this requirement than other tokens. In the PFT stage, we can leverage preference pairs which are similar in content but differ in these instruction-critical tokens to more directly illustrate the causality between the token generation and the successful instruction-following. Second, while the SFT stage can utilize positive examples only, PFT stage can also leverage negative examples to highlight 
the fine difference between a successful instruction-following and a subtle failure.

To this end, we propose a novel technique, Evolutionary Contrastive Distillation (ECD), for generating high-quality, synthetic preference data targeting complex instruction-following. 
Following \citet{xu2023wizardlm, sun2024conifer}, we prompt proprietary LLMs to progressively increase the complexity of instructions. Instead of separating the original example and the evolved example as two independent examples for SFT, however, we connect them as a preference pair for PFT: the proprietary LLM's response on the evolved instruction is considered as a positive example, and its response on the original instruction is considered a negative example. The key observation is that when an instruction is evolved to have a new requirement, the original \emph{good} response for the original instruction is not a good response for the evolved instruction anymore, and therefore can be used as the negative example to the evolved instruction. Since the evolution of instructions is gradual, however, the original response still satisfies most of the requirements of the evolved instruction, and therefore can be used as a "hard" negative for the evolved instruction. See Figure \ref{fig:rlced_data} for a concrete example. 
This method has the desireable properties that (i) it does not rely on LLMs to annotate preferences or generate undesirable responses, which can be unreliable \citep{yang2023rlcd} and (ii) it is effective at creating hard negative examples, which have are crucial in contrastive learning \citep{chopra2005learning, hadsell2006dimensionality, robinson2020contrastive}.

To ensure that each step of instruction evolution provides a high-quality example of subtle nuance in instruction-following, we devise a fine-grained hierarchical taxonomy of evolution operations, which we discuss in Section~\ref{sec:autoevolve}. We demonstrate the data generated from this taxonomy is not only effective at improving complex instruction-following, but can also be combined with previous evolutionary \citep{sun2024conifer} as well as non-evolutionary \citep{cui2023ultrafeedback} methods to yield even stronger results.
Therefore, we believe the proposed taxonomy will serve as a useful resource on its own for future research on complex instruction-following.

We validate the effectiveness of our approach in an extensive set of experiments, benchmarking our models on three recent instruction-following benchmarks: IFEval, FollowBench, and InfoBench. Our approach yields a state-of-the-art 7B model, improving on prior SOTA (state-of-the-art) instruction-following 7B models by 7pp on IFEval and is competitive with popular open-source models at the 70B scale. Furthermore, we develop a recipe to build SOTA instruction-following models that also achieve highly competitive performance on popular conversational quality benchmarks such as MT-Bench and AlpacaEval. Finally, we perform ablations illustrating the advantages of ECD over other synthetic preference data generation methods such as RLAIF and RLCD, and the importance of using contrastive learning (e.g., DPO) instead of SFT to learn from preference pairs.

\section{Related Work}

\paragraph{Instruction-Following.} The complex instruction-following ability of LLMs has received significant attention recently with many works proposing new evaluation benchmarks \citep{zhou2023instruction, jiang2023followbench, qin2024infobench}. 
They consistently find that open source LLMs have significant gaps on following complex instructions compared to proprietary LLMs.
At the same time, there has been relatively less work on developing techniques to \emph{improve} the complex instruction-following ability. Two exceptions include Evol-Instruct \citep{xu2023wizardlm} and Conifer \citep{sun2024conifer}, which prompt LLMs to evolve the complexity of instructions, and apply SFT on the generated data. There are two main differences between these works and ours. First, we focus on generating preference data for PFT instead of demonstration data for SFT.  While SFT data provides only positive feedback to the model on the correct behavior, PFT data provides both positive and negative feedback, which enables teaching the contrast between a successful instruction-following example and a subtle failure example. Second, we propose a fine-grained hierarchical taxonomy of evolution operations to ensure each step of evolution introduces diverse and subtle variations of requirements.


\paragraph{Contrastive Learning.} The effectiveness of contrastive learning \citep{chopra2005learning, hadsell2006dimensionality} has been shown across 
numerous modalities \citep{he2020momentum, chi2020infoxlm, radford2021learning}. 
In the contrastive learning literature, the importance of the quality of negative samples has been well-established \citep{robinson2020contrastive}. Numerous "hard" negative mining techniques have been introduced to improve the quality of negative samples \citep{schroff2015facenet, wu2018sampling, xiong2021approximate}. While many 
alignment methods such as InstructGPT \citep{ouyang2022training} and DPO \citep{rafailov2023direct} leverage negative samples to facilitate learning \citep{tajwar2024preference}, how to obtain sufficiently hard negative samples 
has not been actively studied. \citet{yan2024contrastive} employs hard negative mining
techniques to improve the robustness of the model. Our work instead targets improving
complex instruction-following, and uses LLM prompting to generate negative examples
rather than mining samples from a fixed dataset.



\paragraph{Data Generation Methods for PFT.}
Much attention has been given to how to generate preference data for fine-tuning LLMs. Pioneering works focused on collecting feedback from humans \citep{christiano2017deep, stiennon2020learning, ouyang2022training}. Under this approach, human annotators would observe a set of responses to an instruction and rank them according to their preferences. As human annotation is expensive and difficult to scale, however, follow-up works have proposed alternative synthetic data generation approaches. 

The most popular method of generating PFT data from LLMs
is to sample two independent responses on the same instruction, and then 
ask an LLM to judge which one is of higher quality. This method is often called
RLAIF (Reinforcement Learning from AI Feedback) \citep{bai2022constitutional}. While RLAIF methods have shown encouraging results 
\citep{cui2023ultrafeedback, tunstall2023zephyr, ivison2023camels}, obtaining
high-quality preference annotations from LLMs have been found challenging
\citep{yang2023rlcd, sharma2024critical}. This is because the difference between
two responses on the identical instruction can be minor, and LLM-based preference
annotation can be subject to many confounding factors, such as self-bias and verbosity bias  \citep{zheng2023judging}.

To avoid quality issues from LLM-based preference annotation, 
Reinforcement Learning from Contrastive Distillation (RLCD) \citep{yang2023rlcd} proposes to employ two different prompt templates 
rather than using the same one. One prompt template is designed to elicit desirable responses, and another is designed to elicit undesirable response. 
Such a deliberate design of templates allows RLCD to bypass preference annotation.
As we discuss in Section~\ref{sec:experiments}, however, we find it challenging to prompt-engineer proprietary LLMs to generate undesirable responses.
As proprietary LLMs are already aligned to generate helpful and harmless responses \citep{bai2022training}, very often, their response from the undesirable response template are either as helpful as the response from the desirable response template, 
or trivially unhelpful ("No, I can't answer that question."), limiting their value as a \emph{hard}
negative example.

In contrast to RLHF or RLAIF, ECD does not require preference annotation by neither  humans nor LLMs. This circumvents issues with preference annotation quality. In contrast to easy negative examples from RLCD, negative examples from ECD are hard, as they are helpful responses on a similar but subtly different instruction.

\label{sec:ecd}

\section{Evolutionary Contrastive Distillation}

\begin{figure*}[htbp!]
  \centering
  \includegraphics[width=\textwidth]{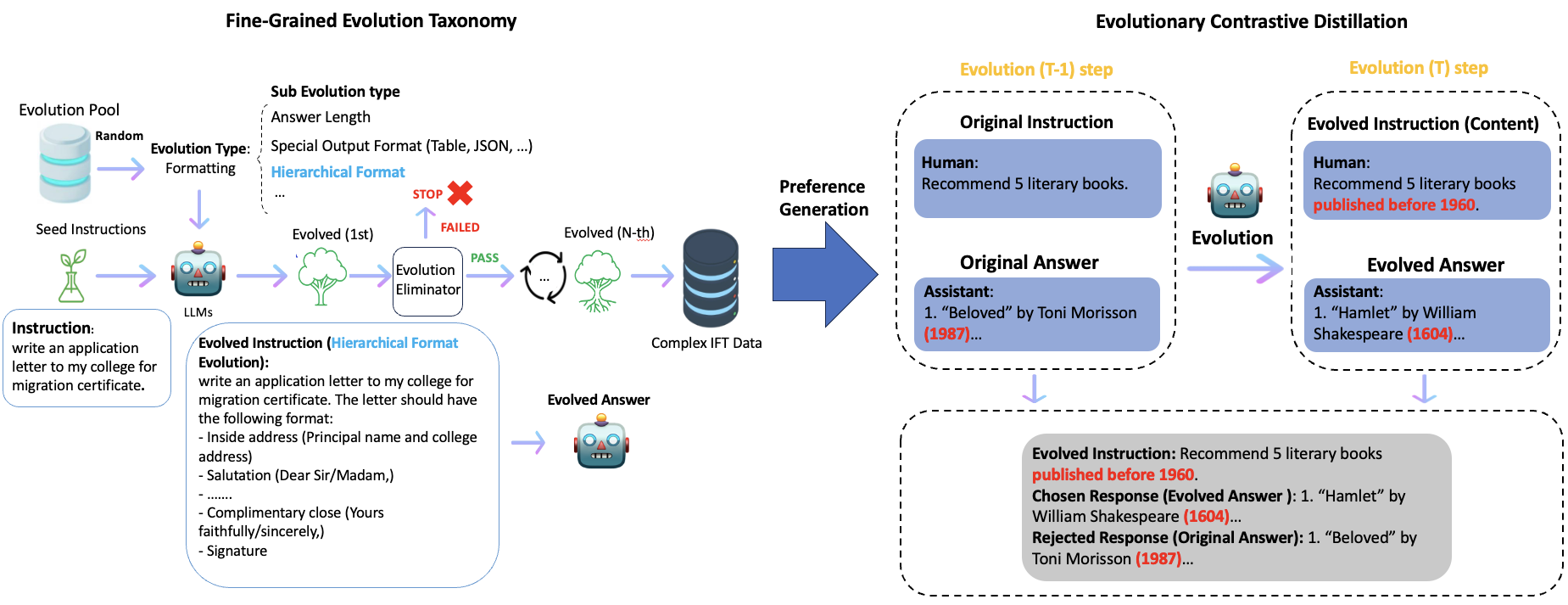}
  \caption{Overview of Evolutionary Contrastive Distillation. The left-hand side depicts the evolutionary process of creating complex instructions and their demonstrations. The right-hand side depicts the process of extracting the preference data from this process.}
  \label{fig:ecd}
\end{figure*}





First, we propose the evolutionary data generation process to synthesize 
preference data for complex instruction-following. 
Then, we discuss how contrastive learning methods shall be employed to
train on this generated data.

\paragraph{Data Generation Framework}

We assume access to a seed set of instructions $\mathcal{I}^{(0)} = \{I^{(0)}_1, \ldots, I^{(0)}_n\}$ and responses on these instructions $\mathcal{R}^{(0)} = \{R^{(0)}_1, \ldots, R^{(0)}_n\}$ where $R^{(0)}_i$ is the response on instruction $I^{(0)}_i$. In our experiments, we use ShareGPT \citep{vicuna2023} as the seed set. 
As instructions which are publicly accessible in scale such as ShareGPT lack the complexity needed for LLM applications \citep{xu2023wizardlm}, we iteratively increase their complexity through $T$ rounds of an evolutionary process, and generate the preference dataset $\mathcal{D}$, which consist of (instruction, positive response, negative response) triples.

At each round $t$, for each instruction $I_i^{(t-1)}$, we run the following:
\begin{itemize}
    \item \textbf{Evolution}: Prompt an LLM to evolve $I_i^{(t-1)}$ into proposal instruction $\tilde{I}_i^{(t)}$.
        The evolution typically increases the complexity of the instruction.
        We discuss types of evolution operations we consider in Section~\ref{sec:autoevolve}.
    \item \textbf{Adaptation}: Prompt an LLM to generate a proposal response $\tilde{R}_i^{(t)}$ on the new      instruction $\tilde{I}_i^{(t)}$.
    \item \textbf{Elimination}: The quality of $\tilde{R}_i^{(t)}$ is checked with another 
        LLM prompt. If the quality is acceptable, the proposal is kept: $(I_i^{(t)}, R_i^{(t)}) \leftarrow (\tilde{I}_i^{(t)}, \tilde{R}_i^{(t)})$. Otherwise,
        $(I_i^{(t)}, R_i^{(t)}) \leftarrow (I_i^{(t-1)}, R_i^{(t-1)})$
    \item \textbf{Contrast}: If the proposal was accepted, add the triple $(I_i^{(t)}, R_i^{(t)}, R_i^{(t-1)})$ into $\mathcal{D}$.
\end{itemize}

The key intuition is that as the original response $R_i^{(t-1)}$ was generated without seeing all of the requirements from the evolved instruction, the evolved response $R_{i}^{(t)}$ is likely better than the original response $R_{i}^{(t-1)}$ for the evolved instruction $I_i^{(t)}$. 
On the other hand, $R_{i}^{(t-1)}$ is still a \emph{hard} negative, as it was a desirable
response for $I_i^{(t-1)}$, and the difference between $I_i^{(t-1)}$ and $I_i^{(t)}$
is often subtle due to the design of evolution operations.

We emphasize that this framework of generating preference data from a evolutionary process
is generic, and can accommodate different definitions of the evolutionary process.
In Section~\ref{sec:experiments}, we show the evolutionary process from \citet{sun2024conifer} can be successfully adopted in this framework.
In order to further improve the quality of the data generated, we propose a 
fine-grained taxonomy of evolution operations in Section~\ref{sec:autoevolve}.





\paragraph{Contrastive Learning} A variety of alignment algorithms can be leveraged to 
fine-tune LLMs with triplets $\mathcal{D}$ generated from the evolutionary process \citep{ouyang2022training, rafailov2023direct, azar2023general, ethayarajh2024kto, hong2024orpo, meng2024simpo}. In this work, we focus on Direct Policy Optimization (DPO) 
\citep{rafailov2023direct} mainly because it is the most well-established in
open source LLMs \citep{tunstall2023zephyr, ivison2023camels}, making our experiments
and checkpoints easily comparable with existing work. We leave it as a future work
to explore the implication of algorithm choice on the complex instruction-following capability.


For completeness, we briefly discuss how DPO is adopted for ECD.
Let us denote $\pi_\theta$ as a language model policy, where $\pi_\theta(R | I)$ denotes the probability of generating the response $R$ conditional on the instruction $I$. We assume that we are given a reference language model $\pi_{\texttt{ref}}$, and use it to initialize $\pi_\theta$. Then, we directly fine-tune the language model $\pi_\theta$ on the preference dataset $\mathcal{D}$ by solving the following optimization problem:

\begin{small}
\begin{align*}
    & \text{argmin}_\theta \mathcal{L}_{\texttt{DPO}}(\mathcal{D} ; \theta) := \\
    &   \mathbb{E}_{I, R^+, R^- \sim \mathcal{D}}\left[ \log\left( \sigma( \frac{\pi_\theta(R^+ | I))}{\pi_{\texttt{ref}}(R^{+}| I)})  - \sigma( \frac{\pi_\theta(R^- | I))}{\pi_{\texttt{ref}}(R^{-}| I)}) \right) \right] .
\end{align*}
\end{small}


Following \citet{rafailov2023direct, rafailov2024r}, the resulting LLM policy $\pi_\theta$
can be associated with the policy which maximizes the reward model learned from 
$\mathcal{D}$. Due to our construction of $\mathcal{D}$, the reward model would
assign higher rewards on responses which meticulously follow complex instructions.


\section{Fine-Grained Evolution Taxonomy}

\label{sec:autoevolve}

In order to ensure the evolutionary process generates diverse variations of instructions,
we define a fine-grained hierarchical taxonomy of evolution operations. 
Each evolution
operation samples a leaf node from this taxonomy to determine the type of operation
and the corresponding prompt template. The top level of the taxonomy consists of five categories:
\begin{itemize}
    \item \textbf{Content}: Add a condition that changes the scope or the specificity of the response.
    \item \textbf{Style}: Control the tone, sentiment, or formality of the response.
    \item \textbf{Format}: Impose a formatting or linguistic requirement.
    \item \textbf{Reasoning}: Add a constraint that requires additional steps of logical or 
    numerical reasoning.
    \item \textbf{Breadth}: Come up with a new instruction that matches the domain, length, and complexity of the original instruction.    
\end{itemize}

Then, for each of the top-level category, we define around five fine-grained evolution types. See Table~\ref{tab:taxonomy} for the full taxonomy of evolution operations. 
While Evol-Instruct \citep{xu2023wizardlm} defines only eleven high-level operation types,
and Confier \citep{sun2024conifer} provides about eleven operation types in in-context examples, our taxonomy defines more fine-grained and hierarchically organized 22 operation types. In Section~\ref{sec:experiments}, we find this taxonomy generates
higher-quality preference data for ECD. 

For illustration, consider the following example of an evolution step:
\begin{enumerate}
    \item The process starts from a seed instruction ``Write an application letter to my college for immigration certificate.''.
    \item \emph{`Format'} is randomly chosen from the top level.
    \item Within the \emph{`Format'} category, \emph{`Hierarchical: Introduce a hierarchical structure which requires an understanding of a hierarchy of tasks and follow it'} is randomly chosen.
    \item LLM is prompted with the prompt corresponding to \emph{`Hierarchical'} type to evolved the original instruction into: ``Write an application letter to my college for immigration certificate. The letter should have the following format: -Date, -Inside address, -Salutation, ...''.
\end{enumerate}
See Figure~\ref{fig:ecd} for a graphical illustration. 

\section{Experiments}

\label{sec:experiments}

\subsection{Models and Datasets}

We conduct our experiments on the popular base model Mistral-7B-v0.1 \citep{jiang2023mistral}. We train our own Instruction Fine-tuned (IFT) model based on the Conifer-7B's SFT recipe  \citep{sun2024conifer}, the SOTA IFT 7B model, mixing in 53k samples from ShareGPT with the Conifer dataset.\footnote{We cannot use the exact same 53k samples from ShareGPT used in the Conifer paper, as this dataset was not made public. We randomly sampled 53k datapoints from \url{https://huggingface.co/datasets/anon8231489123/ShareGPT_Vicuna_unfiltered.} to best mimic it.} We call the resulting IFT dataset Conifer-Mix and the resulting IFT model, Conifer-7B-SFT.\footnote{Since model parameters from \citet{sun2024conifer} were not publicly released, we reproduced the recipe according to the paper to the best of our ability.} Conifer-7B-SFT forms the backbone on which we perform PFT experiments using DPO. For details on training, see Appendix~\ref{app:experiment_details}. 

To test the robustness of our approach, we generate 3 separate ECD datasets: (i) \textbf{ECD-FineGrained}: 30k preference pairs from Fine-Grained Evolutionary Process discussed in Section~\ref{sec:autoevolve},\footnote{We execute this strategy for four rounds of evolution using Claude 2, generating a total of 104,499 preference data from ShareGPT instructions. Then we subsampled 30k.} (ii) \textbf{ECD-Conifer}: ECD data based on the evolutionary process from Conifer \citep{sun2024conifer}\footnote{We removed evolutions with "process feedback"-type, because it does not generate hard negatives needed for ECD.}, and (iii) \textbf{ECD-FineGrained-Conifer}: a concatenation of ECD-FineGrained and ECD-Conifer. We add UltraFeedback \citep{cui2023ultrafeedback} to each of these to improve conversational quality\footnote{We use \url{https://huggingface.co/datasets/HuggingFaceH4/ultrafeedback_binarized?row=1} and remove preference pairs with equal GPT-4 scores.}; see the Appendix~\ref{ssec:uf_ablation} for the ablation on its impact. 

\begin{table*}[t]
\centering
\resizebox{\textwidth}{!}{
\begin{tabular}{l|ccc|cccc|cc|ccc}
\multirow{2}{*}{}  & \multirow{2}{*}{\textbf{Base Model}}  & \multirow{2}{*}{\textbf{IFT Data}}    & \multirow{2}{*}{\textbf{Preference Data}} & \multicolumn{4}{c|}{\textbf{IFEval}}                                            & \multicolumn{2}{c|}{\textbf{FollowBench}} & \multicolumn{3}{c}{\textbf{InfoBench}}               \\ \cline{5-13}
                          & \multicolumn{1}{l}{} & \multicolumn{1}{l}{} & \multicolumn{1}{l|}{}    & \textbf{strict pr} & \textbf{strict in} & \textbf{loose pr} & \textbf{loose in} & \textbf{HSR}        & \textbf{SSR}        & \textbf{Easy}   & \textbf{Hard}   & \textbf{Overall} \\ \hline
\textbf{LLaMa-2-70B-Chat}$^\dagger$ & LLaMa-2              & -                    & -                        & -                  & -                  & \textbf{-}        & \textbf{-}        & 47.5\%              & 55.9\%              & 89.6\%          & 82.1\%          & 84.4\%           \\ \hline
\textbf{Vicuna-13B-v1.5}$^\dagger$  & LLaMa-2              & ShareGPT             & -                        & 43.1\%             & 53.6\%             & 46.6\%            & 58.0\%            & 50.4\%              & 59.5\%              & 85.7\%          & 73.7\%          & 77.3\%           \\ \hline
\textbf{Mistral-7B-Evol-Instruct}$^\dagger$                         & Mistral-7B           & Evol-Instruct        & -                        & 41.4\%             & 51.3\%             & 44.0\%            & 54.4\%            & 40.7\%              & 52.7\%              & 81.0\%          & 73.2\%          & 75.6\%           \\
\textbf{Conifer-7B-SFT}                         & Mistral-7B           & Conifer-Mix          & -                        & 45.1\%             & 58.0\%             & 48.6\%            & 61.2\%            & 49.7\%              & 60.4\%              & 85.9\%          & 80.6\%          & 82.2\%           \\
\textbf{Mistral-7B-ShareGPT-DPO}$^\dagger$                         & Mistral-7B           & ShareGPT             & UltraFeedback            & 43.8\%             & 55.8\%             & 48.2\%            & 59.7\%            & 47.7\%              & 55.9\%              & 86.8\%          & 79.9\%          & 82.0\%           \\
\textbf{Deita-7B-V1.0}$^\dagger$    & Mistral-7B           & Deita-Mix            & UltraFeedback                        & 44.6\%             & 56.6\%             & 51.9\%            & 63.7\%            & 45.7\%              & 54.3\%              & 86.2\%          & 78.6\%          & 80.9\%           \\
\textbf{Conifer-7B-DPO}$^\dagger$   & Mistral-7B           & Conifer-Mix$^\star$          & UltraFeedback            & 48.1\%             & 59.1\%             & 52.3\%            & 63.3\%            & 50.0\%              & 56.2\%              & \textbf{87.5\%} & 80.0\%          & 82.3\%           \\
-                         & Mistral-7B           & Conifer-Mix          & UltraFeedback            & 44.5\%             & 56.5\%             & 52.5\%            & 64.3\%            & 58.7\%              & 65.9\%              & 86.2\%          & 82.4\%          & 83.6\%           \\
\textbf{ECD}                        & Mistral-7B           & Conifer-Mix          & ECD-Conifer              & 47.9\%             & 59.7\%             & 54.9\%            & 66.2\%            & 57.4\%              & 65.4\%              & 86.5\%          & 82.4\%          & 83.7\%           \\
\textbf{ECD}                          & Mistral-7B           & Conifer-Mix          & ECD-FineGrained           & \textbf{52.9\%}    & 63.5\%             & \textbf{59.3\%}   & \textbf{69.8\%}   & \textbf{63.4\%}     & \textbf{70.8\%}     & 84.3\%          & 82.0\%          & 82.7\%           \\
\textbf{ECD}                          & Mistral-7B           & Conifer-Mix          & ECD-FineGrained-Conifer   & 52.3\%             & \textbf{63.8\%}    & 58.8\%            & 68.9\%            & 58.6\%              & 65.4\%              & 85.5\%          & \textbf{85.2\%} & \textbf{85.3\%}  \\ \hline
\end{tabular}
}
\caption{Main results on Instruction Following benchmarks:  IFEval, FollowBench, and InFoBench. Bold-face indicates the best results among the 7B models. $\dagger$ indicates that the results are from the original source. $\star$ denotes the mixture of the Conifer dataset with ShareGPT from the original paper \citep{sun2024conifer}. Note that ``in" abbreviates ``instruction" and ``pr" abreviates ``prompt" in the above table, so for example ``loose in" abbreviates ``loose instruction accuracy". These abbreviations hold for subsequent tables as well.}
\label{tab:main_instruction_following}
\end{table*}

\begin{table*}[]
\centering
{\tiny
\begin{tabular}{l|ccc|c|cc}
\multirow{2}{*}{\textbf{Model}}               &       \multirow{2}{*}{\textbf{Base Model}}              & \multirow{2}{*}{\textbf{IFT Data}} & \multirow{2}{*}{\textbf{Preference Data}}    & \multirow{2}{*}{\textbf{MT-Bench Score}}               & \multicolumn{2}{c}{\textbf{AlpacaEval}}        \\ \cline{6-7}
 &  &     &  &  & \textbf{LC Win-Rate} & \textbf{Average Length} \\ \hline
\textbf{Mistral-7B-Evol-Instruct}$^\dagger$                                   & Mistral-7B          & Evol-Instruct        & -                        & 6.51                    & 9.4\%               & 982                     \\
\textbf{Conifer-7B-SFT}                                     & Mistral-7B          & Conifer-Mix          & -                        & 6.74                    & 10.0\%               & 1002                    \\
\textbf{Deita-7B-v1.0}$^\dagger$              & Mistral-7B          & Deita-Mix            & UltraFeedback            & \textbf{7.55}           & 16.1\%              & 1417                    \\
\textbf{Mistral-7B-ShareGPT-DPO}$^\dagger$                                     & Mistral-7B          & ShareGPT             & UltraFeedback            & 7.1                     & 15.1\%              & 1276                    \\
\textbf{Conifer-DPO-7B}$^\dagger$               & Mistral-7B          & Conifer-Mix$^\star$          & UltraFeedback            & 7.25                    & 17.1\%              & 1253                    \\
\textbf{Zephyr-7B-beta}$^\dagger$               & Mistral-7B          & UltraChat            & UltraFeedback            & 7.34                    & 13.2\%              & 1444                    \\
-                                   & Mistral-7B          & Conifer-Mix          & UltraFeedback            & 7.41                    & 23.3\%              & 1528                    \\
\textbf{ECD}                                   & Mistral-7B          & Conifer-Mix          & ECD-Conifer              & 7.49                    & \textbf{25.2\%}     & 1424                    \\
\textbf{ECD}                                     & Mistral-7B          & Conifer-Mix          & ECD-FineGrained           & 7.35                    & 22.0\%              & 1327                    \\
\textbf{ECD}                                    & Mistral-7B          & Conifer-Mix          & ECD-FineGrained-Conifer   & 7.47                    & 20.6\%              & 1427                    \\ \hline
\end{tabular}
}
\caption{Main results on conversational quality benchmarks: MT-Bench and AlpacaEval. $\dagger$ indicates that the results are from the original paper. $\star$ denotes the mixture of the Conifer dataset with ShareGPT from the original source. $\star$ denotes the mixture of the Conifer dataset with ShareGPT from the original paper \citep{sun2024conifer}.}
\label{tab:main_conv_quality}
\end{table*}

\subsection{Evaluation}

Our primary goal is to improve the complex instruction-following capability of LLMs, which we measure with the following three benchmarks:
\begin{itemize}
    \item \textbf{IFEval} is a popular benchmark for instruction-following that measures the ability of LLMs to follow programatically checkable instructions such as “give a response that is more than 400 words" \citep{zhou2023instruction}. It uses metrics such as prompt-level accuracy and instruction-level accuracy with a strict version that interprets the requirements very precisely while a loose version gives some leeway.
    \item \textbf{FollowBench} is another instruction-following benchmark that uses GPT-4 to measure the ability of a model to follow fine-grained constraints across 5 different difficulties and types (Format, Content, Style, Situation, and Example) \citep{jiang2023followbench}. This benchmark employs two  metrics: hard satisfaction rate (HSR) and soft satisfaction rate (SSR). HSR quantifies the average frequency at which all the requirements or constraints are completely met. On the other hand, SSR calculates the average degree to which individual constraints are satisfied across all the given instructions.
    \item \textbf{InfoBench} evaluates the instruction-following of LLMs by breaking down instructions into a set of fine-grained criteria and asks GPT-4 to evaluate the extent to which a model meets the criteria \citep{qin2024infobench}. 
\end{itemize}
Since conversational quality is also important in LLM applications, we also evaluate on the following benchmarks:
\begin{itemize}
    \item \textbf{MTBench} is a multi-turn benchmark that measures the conversational quality of a model. It uses GPT-4 to rate the quality of a model's answers across two turns on a scale of 1-10 \citep{zheng2023judging}.
    \item \textbf{AlpacaEval} is a single-turn benchmark that measures helpfulness. It uses GPT-4-Turbo to compute the win-rate against a reference model. We use the default reference model, GPT-4-Turbo, and the Length-Controlled win-rate, which has a correlation of 0.98 with ChatBot Arena \citep{dubois2024alpacafarm, dubois2024length}.
\end{itemize}


We benchmark our ECD models against large open-source scale models such as LLaMa-2-70B-Chat \citep{touvron2023llama} and Vicuna-13B-v1.5 \citep{vicuna2023}, and strong 7B models like Conifer-DPO-7B \citep{sun2024conifer}, Deita-7B-v1.0 \citep{liu2023makes}, Mistral-7B-Evol-Instruct \citep{xu2023wizardlm}, Mistral-7B-ShareGPT-DPO \citep{sun2024conifer}, and Zephyr-7B-beta \citep{tunstall2023zephyr}.\footnote{For the models Mistral-7B-Evol-Instruct and Mistral-7B-ShareGPT-DPO, we report the evaluation numbers given in \cite{sun2024conifer}.}

\subsection{Results}

Our ECD models achieve SOTA performance at the 7B scale for complex instruction-following. For example, consider our ECD model trained on ECD-FineGrained-Conifer. Table~\ref{tab:main_instruction_following} shows that it outperforms Conifer-7B-DPO, the latest 7B SOTA in instruction-following, on each metric in IFEval by a substantial margin, improving loose prompt accuracy from 52.3\% to 59.3\% and loose instruction accuracy from 63.3\% to 69.8\%, and achieves similar improvements on FollowBench and InfoBench. Similarly, the ECD on ECD-Finegrained-Conifer improves over its initialization Conifer-7B-SFT, improving for example on loose prompt accuracy by over 10pp and even shows competitive performance with LLaMa-2-70B-Chat. ECD-FineGrained-Conifer achieves particularly strong performance on InfoBench Hard, indicating the strength of ECD in improving instruction-following for particularly difficult instructions. While here we discuss specifically ECD-FineGrained-Conifer, these trends uphold across our three ECD data mixtures, ECD-FineGrained, ECD-Conifer, and ECD-FineGrained-Conifer, indicating the robustness of our approach.


Our ECD models also achieve strong performance on the conversational quality benchmarks: MT-Bench and AlpacaEval. For example, consider ECD-Conifer. It achieves an MT-Bench score of 7.49 and and a length-controlled AlpacaEval score of 25.2\%. These are large improvements in comparison to the initialized IFT model Conifer-7B-SFT and the prior SOTA in instruction-following 7B models, Conifer-7B-DPO. 
As this trend holds across all three ECD models, our data mixture recipe consistently produces SOTA 7B models for complex instruction-following while maintaining strong conversational quality.



\begin{table*}[t]
\centering
{\tiny
\begin{tabular}{ccc|cccc|cc|ccc}
\multirow{2}{*}{\textbf{Base Model}} & \multirow{2}{*}{\textbf{IFT Data}} & \multirow{2}{*}{\textbf{Preference Data}}    & \multicolumn{4}{c|}{\textbf{IFEval}}                                            & \multicolumn{2}{c|}{\textbf{FollowBench}} & \multicolumn{3}{c}{\textbf{InfoBench}}               \\ \cline{4-12}
                    &                   &                             & \textbf{strict pr} & \textbf{strict in} & \textbf{loose pr} & \textbf{loose in} & \textbf{HSR}        & \textbf{SSR}        & \textbf{Easy}   & \textbf{Hard}   & \textbf{Overall} \\ \hline
Mistral-7B          & Conifer-Mix       & \textbf{}                   & 45.1\%             & 58.0\%             & 48.6\%            & 61.2\%            & 49.7\%              & 60.4\%              & 85.9\%          & 80.6\%          & 82.2\%           \\
Mistral-7B          & Conifer-Mix       & ECD-FineGrained-Pure         & 58.4\%             & 67.5\%             & 62.7\%            & 71.8\%            & 57.0\%              & 65.2\%              & 84.3\%          & 79.2\%          & 80.8\%           \\
Mistral-7B          & Conifer-Mix       & ECD-Conifer-Pure            & 53.0\%             & 64.3\%             & 57.7\%            & 69.1\%            & \textbf{57.8\%}     & 64.4\%              & \textbf{88.4\%} & \textbf{83.8\%} & \textbf{85.2\%}           \\
Mistral-7B          & Conifer-Mix       & ECD-FineGrained-Conifer-Pure & \textbf{63.8\%}    & \textbf{72.5\%}    & \textbf{67.5\%}   & \textbf{76.1\%}   & 57.5\%              & \textbf{65.9\%}     & 86.2\%          & 80.7\%          & 82.4\%  \\
Mistral-7B          & Conifer-Mix       & ShareGPT-RLAIF              & 35.9\%             & 48.4\%             & 47.0\%            & 59.0\%            & 56.0\%              & 65.4\%              & 85.2\%          & 79.3\%          & 81.1\%           \\
Mistral-7B          & Conifer-Mix       & ShareGPT-RLCD               & 45.3\%             & 56.7\%             & 49.5\%            & 60.1\%            & 50.7\%              & 62.2\%              & 85.2\%          & 79.6\%          & 81.3\%           \\ \hline
\end{tabular}
}
\caption{Main table on Instruction Following benchmarks comparing ECD, RLCD, and RLAIF.}
\label{tab:instruction_following_method_comparision}
\end{table*}

\begin{table*}[t]
\centering
{\small
\begin{tabular}{ccc|c|cc}
\multirow{2}{*}{\textbf{Base Model}} & \multirow{2}{*}{\textbf{IFT Data}} & \multirow{2}{*}{\textbf{Preference Data}}    & \multirow{2}{*}{\textbf{MT-Bench Score}}               & \multicolumn{2}{c}{\textbf{AlpacaEval}}        \\ \cline{5-6}
\textbf{}           & \textbf{}         & \textbf{}                   &  & \textbf{LC Win-Rate} & \textbf{Average Length} \\ \hline
Mistral-7B          & Conifer-Mix       & \textbf{}                   & 6.74                    & 10.0\%               & 1002                    \\
Mistral-7B          & Conifer-Mix       & ECD-FineGrained-Pure         & 6.61                    & 10.4\%              & 875                     \\
Mistral-7B          & Conifer-Mix       & ECD-Conifer-Pure            & 6.24                  & \textbf{14.5\%}     & 973                     \\
Mistral-7B          & Conifer-Mix       & ECD-FineGrained-Conifer-Pure & 6.61                    & 10.4\%              & 834                     \\
Mistral-7B          & Conifer-Mix       & ShareGPT-RLAIF              & \textbf{6.87}           & 10.2\%              & 1300                    \\
Mistral-7B          & Conifer-Mix       & ShareGPT-RLCD               & 6.81                    & 10.4\%                 & 1075                    \\ \hline
\end{tabular}
}
\caption{Main table on conversational quality benchmarks comparing ECD, RLCD, and RLAIF.}
\label{tab:conv_quality_method_comparison}
\end{table*}

\begin{table*}[th!]
\centering
{\scriptsize
\begin{tabular}{lc|cccc|cc|ccc}
\multirow{2}{*}{\textbf{Model}} & \multirow{2}{*}{\textbf{Final Stage}} & \multicolumn{4}{c|}{\textbf{IFEval}}                                            & \multicolumn{2}{c|}{\textbf{FollowBench}} & \multicolumn{3}{c}{\textbf{InfoBench}}                      \\ \cline{3-11}
\multicolumn{1}{c}{}               &                      & \textbf{strict pr} & \textbf{strict in} & \textbf{loose pr} & \textbf{loose in} & \textbf{HSR}        & \textbf{SSR}        & \textbf{Easy} & \textbf{Hard} & \textbf{Overall}            \\
\hline
\textbf{Conifer-7B-SFT}                     & SFT                  & 45.1\%             & 58.0\%             & 48.6\%            & 61.2\%            & 49.7\%              & 60.4\%              & \textbf{85.9\%}        & 80.6\%        & 82.2\%                      \\
\textbf{ECD-FineGrained-SFT}                 & SFT                  & 44.5\%             & 56.2\%             & 49.0\%            & 60.0\%            & 42.0\%              & 57.2\%              & 80.7\%        & 74.2\%        & \multicolumn{1}{r}{76.2\%} \\
\textbf{ECD-FineGrained}                     & DPO                  & \textbf{52.9\%}    & \textbf{63.5\%}    & \textbf{59.3\%}   & \textbf{69.8\%}   & \textbf{58.6\%}     & \textbf{70.8\%}     & 84.3\%        & \textbf{82.0\%}        & \textbf{82.7\%}                      \\ \hline
\end{tabular}
}
\caption{Main table on Instruction Following benchmarks, IFEval and FollowBench, comparing DPO and SFT.}
\label{tab:instruction_following_sft_dpo}
\end{table*}

\textbf{Among ECD, RLAIF, and RLCD, which is the most effective technique for improving complex instruction-following?} We also investigated how ECD compares against RLAIF and RLCD for improving instruction-following. To this end, we generated RLAIF and RLCD data\footnote{We used Claude 2 to be consistent with ECD-FineGrained.} on top of ShareGPT prompts and refer to these as ShareGPT-RLAIF and ShareGPT-RLCD, respectively. Prompts used in the generation can be found in the Appendix~\ref{sec:prompts}.

 In order to perform a clean ablation, we used three further data mixtures which do not mix UltraFeedback in: (i) ECD-FineGrained-Pure, all 104,499 preference pairs from the ECD version of the FineGrained synthetic data generation approach, (ii) ECD-Conifer-Pure, the ECD version of the Conifer dataset, and (iii) ECD-FineGrained-Conifer-Pure, a concatenation of ECD-FineGrained and ECD-Conifer-Pure. Since ShareGPT-RLAIF, ShareGPT-RLCD, and ECD-FineGrained-Pure all use the same ShareGPT instructions as seeds and Claude 2 as the teacher LLM, we can directly compare the performance of the models trained on these three data mixtures to assess the effectiveness of ECD, RLAIF, and RLCD.

 While ECD methods show a robust ability to improve instruction-following, RLAIF and RLCD show uneven performance. Table \ref{tab:instruction_following_method_comparision} shows that ECD improves over the IFT initialization on the instruction-following benchmarks. For example, the ECD-FineGrained-Pure mixture achieves a loose prompt accuracy of 62.7\% and the ECD-FineGrained-Conifer-Pure mixture achieves a loose prompt accuracy of 67.5\% in comparison to 48.6\% for its initialization Conifer-7B-SFT. On the other hand, RLAIF and RLCD show no improvement on IFEval and InfoBench while RLAIF only shows some marginal improvement over its initialization Conifer-7B-SFT on FollowBench.
 

On the other hand, for conversational quality, we observe in Table \ref{tab:conv_quality_method_comparison} that ECD by itself yields mostly no improvement, while RLCD and RLAIF show slight improvements. This finding highlights that the present instantiation of ECD is primarily a method for improving instruction-following.

\textbf{What is the impact of using SFT instead of DPO?} We investigated the importance of DPO for the instruction-following ability of our models. In particular, we performed an epoch of SFT instead of DPO on positive respones from ECD-FineGrained, denoted  ECD-FineGrained-SFT. Table~\ref{tab:instruction_following_sft_dpo} shows that SFT on ECD-FineGrained underperforms its initialization Conifer-7B-SFT, while DPO on the same data makes strong improvements. This indicates PFT with contrastive learning is a more effective method for improving complex instruction-following, compared to SFT.

\section{Conclusion}
\label{sec:conclusion}

In this paper, we proposed a novel approach to generate high-quality synthetic preference data for complex instruction-following, Evolutionary Contrastive Distillation. Using this approach, we trained models at the 7B scale that achieved state-of-the-art performance in instruction-following as measured by benchmarks such as IFEval, FollowBench, and InfoBench and achieved competitive performance on conversational quality benchmarks like AlpacaEval and MT-Bench. For example, one of our checkpoints improves over the prior SOTA, Conifer-7B-DPO, at the 7B scale on IFEval loose prompt accuracy by 7pp while also achieving a score of 7.35 on MT-Bench and a length-controlled win-rate against GPT-4-turbo of 22\%. 

Due to the generality of our approach, we believe that there are opportunities to apply it to many other domains beyond complex instruction-following. Indeed, recent research has already produced evolved IFT data for mathematics and coding \citep{luo2023wizardmath, luo2023wizardcoder} with impressive results and our approach offers a way to convert these datasets into preference data so that models can learn from both positive and negative feedback. As part of our future work, we plan to explore these other domains.

\section{Limitations}

In this work, we focused on improving complex instruction-following capability.
However, we envision that ECD can also be useful at improving other LLM capabilities,
such as tool usage \citep{schick2023toolformer}, reasoning \citep{talmor2018commonsenseqa}, math \citep{cobbe2021training}, etc.
Broadening the definition of our evolutionary process to target a broader set of
capabilities is left as a future work.

Also, in this work, we focused on having a dependency on a teacher LLM to evolve instructions and generate responses.
Therefore, the resulting model is likely to inherit various types of bias the teacher LLM has.
However, it is conceivable the teacher model in ECD could be replaced with the student
model itself. Such self-improvement \citep{yuan2024selfrewarding} will remove dependency
on teacher models, and open up an opportunity to surpass them. Future work is required to determine whether it is possible to remove the dependency on a strong teacher LLM. 

\section{Ethical Considerations}

The focus of our work is on improving the complex instruction-following capability of LLMs, a fundamental capability. 
The ability to faithfully execute instructions from humans will
reduce the risk of LLMs undertaking unintended actions, promoting safer uses of LLMs.
On the flip side, this can increase the risk of malicious actors
using LLMs towards pernicious ends. A fruitful direction for future research is to continue using alignment to improve complex instruction-following while enhancing the model's capability to refrain from executing harmful tasks.  


\newpage

\null

\appendix

\section{Evolution Details}

\subsection{Full Taxonomy}
See Table~\ref{tab:taxonomy} for the full taxonomy of evolution operations. We also show the prompts for fine-grained evolution in Section~\ref{ssec:fine_grained_evolution}. Most of the evolution operations are designed to introduce gradual, nuanced changes to the original instruction. Examples of these evolutions can be found in Table~\ref{tab:evolution_examples}. Additionally, since the proposed method leverages DPO, it remains robust to occasional poor examples in the negative responses. For such "easy" negative responses, where the margin between positive and negative responses is large, the gradient magnitude will be small. This highlights another advantage of Preference Fine-Tuning over Supervised Fine-Tuning.

\subsection{Evolution Elimination}
We utilized an evolution eliminator employing heuristic methods such as instruction length variation and deduplication to determine if the evolved instructions and answers are both valid. If unsuccessful, halt the process; otherwise, proceed to the next round of evolution. This precaution is necessary because LLMs like Claude2 may also produce errors, such as omitting code snippets/tables from the original instruction, or generating duplicate instructions.


\section{Additional Details}
\label{app:experiment_details}

\subsection{Training Details}

We leverage the widely adopted 'The Alignment Handbook' \citep{tunstall2023zephyr} repository, released by HuggingFaceH4, for fine-tuning. For all our experiments, we use a machine with 8 NVIDIA A100 80GB GPUs. For SFT, we train for 4 epochs with a learning rate of $2e{-5}$ and warm-up ratio of $0.1$ with per device batch size 16 and gradient accumulation steps 4. For DPO training, we train for 1 epoch a per device batch size of 8 and a gradient accumulation of 2. We use a learning rate of $5e{-7}$ and a warm-up ratio of $0.1$.

\begin{table*}[t]
    \centering
    \resizebox{\textwidth}{!}{
\begin{tabular}{l|l}
\multicolumn{1}{c|}{\multirow{2}{*}{\textbf{Category}}} & \multicolumn{1}{c}{\multirow{2}{*}{\textbf{Fine-grained type}}}                                                                    \\
\multicolumn{1}{c|}{}                                   & \multicolumn{1}{c}{}                                                                                                               \\ \hline
\multirow{6}{*}{\textbf{Content}}                       & Add a Subtask or Another Related Question.                                                                                         \\
                                                        & Set a Higher standard: Raise the bar for what's considered acceptable or successful.                                               \\
                                                        & Set a Higher Standard: Raise the bar for what's considered acceptable or successful.                                               \\
                                                        & Limit resources: Restrict the number or type of resources that can be used                                                         \\
                                                        & Add a criterion: Mandate a new feature to be included.                                                                             \\
                                                        & Sequencing: Dictate the order in which steps or actions should be taken.                                                           \\ \hline
\multirow{5}{*}{\textbf{Style}}                         & Tone and Emotion: Specify the desired emotional tone for the response.                                                             \\
                                                        & Ask to mimic a specific author's writing style.                                                                                    \\
                                                        & Contradiction: Ask to provide a response that contradicts the previous statement or take a stance opposite to its prior response.  \\
                                                        & Ambiguity: Create responses with intentional ambiguity or double meanings.                                                         \\
                                                        & Humor or Satire: Request to be humorous or satirical, generating jokes or witty remarks.                                           \\ \hline
\multirow{7}{*}{\textbf{Format}}                        & Length: Imposing constraints on the length of individual words, sentences, or paragraphs.                                          \\
                                                        & Hierarchical: Introduce a hierarchical structure which requires an understanding of a hierarchy of tasks and follow it.            \\
                                                        & Special output format: Use data formats like table, json, HTML, LaTeX, etc.                                                        \\
                                                        & Morphological constraints: Use or avoid specific morphemes.                                                                        \\
                                                        & Multi-lingual Constraints: Respond in multiple languages or switch between languages.                                              \\
                                                        & Incorporate literary devices: Introduce specification(s) of metaphor, simile, alliteration, irony, symbolism, foreshadowing, etc.. \\
                                                        & Grammatical structure: Strictly follow a particular grammatical structure.                                                         \\ \hline
\multirow{3}{*}{\textbf{Reasoning}}                     & Explicitly request multiple-step or chain-of-thought reasoning.                                                                    \\
                                                        & Add some numeric reasoning steps.                                                                                                  \\
                                                        & Add some commonsense reasoning steps.                                                                                              \\ \hline
\textbf{Breadth}                                        & Come up with a new instruction that matches the domain, length, and complexity of the original instruction                         \\ \hline
\end{tabular}
}
    \caption{Full Taxonomy of Evolution Operations}
    \label{tab:taxonomy}
\end{table*}

\begin{table*}[t]
    \centering
    \resizebox{1.5\columnwidth}{!}{
\begin{tabular}{l|p{5cm}|p{5cm}}
\textbf{Category}  & \multicolumn{1}{c|}{\textbf{Original}}                                                                                                                                                   & \multicolumn{1}{c}{\textbf{Evolve (iter+1)}}                                                                                                                                                                                                                                                                        \\ \hline
\textbf{Content}   & Recommend 5 books to me.                                                                                                                                                                 & Recommend 5 \textbf{historic} books to me.                                                                                                                                                                                                                                                                                   \\ \hline
\textbf{Style}     & Some days ago I run for the first time 21 kilometers. It was the longest I ever run. I'm very happy about it and I've been meaning to write some reflexitions about it. Can you help me? & Some days ago I run for the first time 21 kilometers. It was the longest I ever run. I'm very happy about it and I've been meaning to write some reflexitions about it. Can you help me \textbf{reflect on this achievement in an inspiring tone that might motivate others to challenge themselves physically as well?}     \\ \hline
\textbf{Format}    & why are vitamins named Vitamin A, B, etc?                                                                                                                                                & why are vitamins named Vitamin A, B, etc? \textbf{Please respond in a concise paragraph format with a length limit of 3-5 sentences.}                                                                                                                                                                                        \\ \hline
\textbf{Reasoning} & make a table of all us presidents since 1924. add columns for wifes height, main campaign issue, number of children, and whether or not they met with the dalai lama                     & make a table of all us presidents since 1924. add columns for wifes height, main campaign issue, number of children, and whether or not they met with the dalai lama. \textbf{Also add a column for the president's age at inauguration and calculate their age by subtracting their birth year from the inauguration year.} \\ \hline
\textbf{Breadth}   & can you give me an example of different consolidation methods of an organizational LCA?                                                                                                  & \textbf{Could you provide an example of the different methods a multinational corporation might use to consolidate the financial statements of its various subsidiaries located around the world?}                                                                                                                  \\ \hline
\end{tabular}

}
    \caption{Examples of finegrained evolutions.}
    \label{tab:evolution_examples}
\end{table*}

\subsection{Ablation Study on Removing UltraFeedback from Data Mixture}
\label{ssec:uf_ablation}

In this section, we conduct an ablation study on removing UltraFeedback from the ECD-Conifer data mixture. Again, we use the Conifer-7B-SFT model as the IFT initialization. We compare two checkpoints ECD-Conifer and ECD-Conifer-Pure. Whereas ECD-Conifer consists of both the ECD version of Conifer and Ultrafeedback to optimize both instruction-following and conversational quality, ECD-Conifer-Pure only removes UltraFeedback. Table \ref{tab:ultrafeedback_ablation_instruction_following} depicts the results for instruction-following and Table \ref{tab:ultrafeedback_ablation_cq} depicts the results on conversational quality. On instruction-following, we see that ECD-Conifer-Pure tends to outperform ECD-Conifer, with particularly strong performance on IFEval. For example, it improves the strict prompt accuracy by 5.1\%. On the other hand, for conversational quality, ECD-Conifer improves on ECD-Conifer-Pure with a much improve MT-Bench score and LC Win-Rate, indicating the usefulness of UltraFeedback for conversational quality.

\begin{table*}
\centering
\resizebox{\textwidth}{!}{
\begin{tabular}{ccc|cccc|cc|ccc}
\textbf{Base Model}  & \textbf{IFT Data}    & \textbf{Preference Data} & \multicolumn{4}{c|}{\textbf{IFEval}}                                            & \multicolumn{2}{c|}{\textbf{FollowBench}} & \multicolumn{3}{c}{\textbf{InfoBench}}               \\ \hline
\multicolumn{1}{l}{} & \multicolumn{1}{l}{} & \multicolumn{1}{l|}{}    & \textbf{strict pr} & \textbf{strict in} & \textbf{loose pr} & \textbf{loose in} & \textbf{HSR}        & \textbf{SSR}        & \textbf{Easy}   & \textbf{Hard}   & \textbf{Overall} \\
Mistral-7B           & Conifer-Mix          & ECD-Conifer              & 47.9\%             & 59.7\%             & 54.9\%            & 66.2\%            & 57.4\%              & \textbf{65.4\%}     & 86.5\%          & 82.4\%          & 83.7\%           \\
Mistral-7B           & Conifer-Mix          & ECD-Conifer-Pure         & \textbf{53.0\%}    & \textbf{64.3\%}    & \textbf{57.7\%}   & \textbf{69.1\%}   & \textbf{57.8\%}     & 64.4\%              & \textbf{88.4\%} & \textbf{83.8\%} & \textbf{85.2\%} 
\end{tabular}
}
\caption{Instruction-Following Benchmarks for ablation of removing UltraFeedback from ECD-Conifer-Pure.}
\label{tab:ultrafeedback_ablation_instruction_following}
\end{table*}

\begin{table*}
\centering
{\small
\begin{tabular}{ccc|c|cc}
                    & \multicolumn{1}{l}{} & \multicolumn{1}{l|}{}    & \textbf{}               & \multicolumn{2}{c}{\textbf{AlpacaEval}}        \\ \hline
\textbf{Base Model} & \textbf{IFT Data}    & \textbf{Preference Data} & \textbf{MT-Bench Score} & \textbf{LC Win-Rate} & \textbf{Average Length} \\
Mistral-7B          & Conifer-Mix          & ECD-Conifer              & \textbf{7.49}           & \textbf{25.21\%}     & 1424                    \\
Mistral-7B          & Conifer-Mix          & ECD-Conifer-Pure         & 6.2375                  & 14.51\%              & 973                    
\end{tabular}
}
\caption{Response Quality Benchmarks for ablation of removing UltraFeedback from ECD-Conifer-Pure.F.}
\label{tab:ultrafeedback_ablation_cq}
\end{table*}

\newpage

\onecolumn

\section{Prompts Used}

\label{sec:prompts}

\subsection{Prompt for Fine-grained Evolution}
\label{ssec:fine_grained_evolution}

\subsubsection{Content Evolution}

\label{ssec:content_prompt}

\begin{small}
\begin{verbatim}
You are an Instruction Rewriting Expert. You need to rewrite #Given Instruction# based on 
#Rewriting Requirement#, in order to obtain a #Rewritten Instruction#. 
Basically, #Rewritten Instruction# should adhere to the following guidelines:
1. #Rewritten Instruction# must be reasonable and must be understood and responded by humans.
2. You should try your best not to make the #Rewritten Instruction# become verbose, 
#Rewritten Instruction# can only add 10 to 20 words into #Given Instruction#.

#Given Instruction#
{given_instruction}

#Rewriting Requirement#
Please add one proper content constraint to the #Given Instruction#. 
The content constraints include but are not limited to: 
1. Add a Subtask or Another Related Question.
2. Narrow Down the Topic: Instead of a general theme or topic, provide a more specific subset.
3. Set a Higher Standard: Raise the bar for what's considered acceptable or successful.
4. Limit Resources: Restrict the number or type of resources someone can use.
5. Introduce Specific Criteria: Mandate particular components or features that must be included.
6. Specifying Sequence: Dictate the order in which certain steps or actions should be taken.

Please start with the sentence "Here is the new instruction:" in #Rewritten Instruction#.
Please don't add anything related to the #Rewriting Requirement# in the #Rewritten Instruction#.
If #Given Instruction# contains no-text parts such as table and code examples
, #Rewritten Instruction# should also keep them.

#Rewritten Instruction#
\end{verbatim}
\end{small}

\subsubsection{Format Evolution}

\label{ssec:format_prompt}

\begin{small}
\begin{verbatim}
You are an Instruction Rewriting Expert. You need to rewrite #Given Instruction#
based on #Rewriting Requirement#, in order to obtain a #Rewritten Instruction#.
Basically, #Rewritten Instruction# should adhere to the following guidelines:
1. #Rewritten Instruction# must be reasonable and must be understood and responded by humans.
2. You should try your best not to make the #Rewritten Instruction# become verbose, 
#Rewritten Instruction# can only add 10 to 20 words into #Given Instruction#.

#Given Instruction#
{given_instruction}

#Rewriting Requirement#
Please add one proper format constraint that #Given Instruction#
does not have. The format constraints include but are not limited to:
1. Length: Imposing constraints on the length of individual words, sentences, or paragraphs.
2. Hierarchical Instructions: Providing instructions that have a hierarchical structure, where the AI
needs to understand and follow a hierarchy of tasks to construct a response.
3. Special Output Format: Asking the AI to respond by using data format like table, json, HTML, LaTeX, etc.
4. Morphological Constraints: Asking the AI to avoid or use specific morphemes.
5. Multi-lingual Constraints: Asking the AI to respond 
in multiple languages or switch between languages according to complex patterns.
6. Incorporation of Specific Literary Devices: 
Requiring the inclusion of specific, and perhaps numerous, literary devices.
7. Following a Specific Grammatical Structure: 
Requiring the AI to create responses that strictly follow a particular grammatical structure.

Please start with the sentence "Here is the new instruction:" in #Rewritten Instruction#.
Please don't add anything related to the #Rewriting Requirement# in the #Rewritten Instruction#.
If #Given Instruction# contains no-text parts such as table and code examples
, #Rewritten Instruction# should also keep them.

#Rewritten Instruction#
\end{verbatim}
\end{small}

\subsubsection{Style Evolution}

\label{ssec:style_prompt}

\begin{small}
\begin{verbatim}
You are an Instruction Rewriting Expert. You need to rewrite #Given Instruction# based on 
#Rewriting Requirement#, in order to obtain a #Rewritten Instruction#. 
Basically, #Rewritten Instruction# should adhere to the following guidelines:
1. #Rewritten Instruction# must be reasonable and must be understood and responded by humans.
2. You should try your best not to make the #Rewritten Instruction# become verbose, 
#Rewritten Instruction# can only add 10 to 20 words into #Given Instruction#.

#Given Instruction#
{given_instruction}

#Rewriting Requirement#
Please add one proper style constraint that #Given Instruction# 
does not have. The style constraints include but are not limited to:
1. Tone and Emotion: Specify the desired emotional tone for the response.
2. Writing Style: Ask the AI to mimic a specific author's writing style.
3. Contradiction: Ask the AI to provide a response that contradicts the previous
statement or take a stance opposite to its prior response. 
4. Ambiguity: Instruct the AI to create responses with intentional ambiguity or double meanings.
5. Humor or Satire: Request that the response be humorous
or satirical, requiring the AI to generate jokes or witty remarks.

Please start with the sentence "Here is the new instruction:" in #Rewritten Instruction#.
Please don't add anything related to the #Rewriting Requirement# in the #Rewritten Instruction#.
If #Given Instruction# contains no-text parts such as table and code examples
, #Rewritten Instruction# should also keep them.

#Rewritten Instruction#
\end{verbatim}
\end{small}

\subsubsection{Breadth Evolution}

\label{ssec:breadth_prompt}

\begin{small}
\begin{verbatim}
You are an Instruction Creator Expert. You need to draw inspiration from the #Given Instruction# 
to create a brand new #Created Instruction# based on #Creation Requirement#. 

#Given Instruction#
{given_instruction}

#Creation Requirement#
1. #Created Instruction# must be reasonable and must be understood and responded by humans.
2. #Created Instruction# should belong to the same domain as the #Given Instruction# 
but be even more rare.
3. The LENGTH and complexity of the #Created Instruction# should be similar to that of the 
#Given Instruction#.
4. '#Given Instruction#', '#Created Instruction#', 'given instruction' and 'created instruction' are not
allowed to appear in #Created Instruction#
5. #Created Instruction# must be self-contained.

Please start with the sentence "Here is the new instruction:" in #Created Instruction#.\n
Please don't add anything related to the #Creation Requirement# in the #Created Instruction#.

#Created Instruction#
\end{verbatim}
\end{small}

\subsubsection{Reasoning Evolution}

\label{ssec:reasoning_prompt}

\begin{small}
\begin{verbatim}
You are an Instruction Rewriting Expert. You need to rewrite #Given Instruction# based on 
#Rewriting Requirement#, in order to obtain a #Rewritten Instruction#. Basically, 
#Rewritten Instruction# should adhere to the following guidelines:
1. #Rewritten Instruction# must be reasonable and must be understood and responded by humans.
2. You should try your best not to make the #Rewritten Instruction# become verbose, 
#Rewritten Instruction# can only add 10 to 20 words into #Given Instruction#.

#Given Instruction#
{given_instruction}

#Rewriting Requirement#
Please add one proper reasoning constraint that #Given Instruction# does not have. The reasoning 
constraints include but are not limited to:
1. Explicitly request multiple-step or chain-of-thought reasoning. 
2. Add some numeric reasoning steps.
3. Add some commonsense reasoning steps.

Please start with the sentence "Here is the new instruction:" in #Rewritten Instruction#.
Please don't add anything related to the #Rewriting Requirement# in the #Rewritten Instruction#.
If #Given Instruction# contains no-text parts such as table and code examples
, #Rewritten Instruction# should also keep them.

#Rewritten Instruction#
\end{verbatim}
\end{small}

\subsection{Prompt for RLAIF}

\label{ssec:rlaif_prompt}

We adopted the RLAIF prompt used from \citep{yang2023rlcd}:

\begin{verbatim}
Consider the following conversation between a human and an assistant:

$instruction

Please choose the response that is more helpful.

Options:

(A) $answer1
(B) $answer2

The answer is: (
\end{verbatim}

\subsection{Prompts for RLCD}

\label{ssec:rlcd_prompt}

We adopted the RLCD prompt used from \citep{yang2023rlcd} for helpful template:
\begin{verbatim}
Human: $instruction

Assistant: (giving a helpful response)
\end{verbatim}
As well as unhelpful template:
\begin{verbatim}
Human: $instruction

Assistant: (giving an unhelpful response)
\end{verbatim}


\cleardoublepage

\section{Licences and Terms of Service Compliance}

Below, we give all the artifacts that we used and their respective licenses:

\begin{itemize}
    \item Alignment Handbook: Apache-2.0
    \item Mistral 7B: Apache-2.0
    \item IFEval: Eclipse Public License - v 2.0
    \item FollowBench: Apache 2.0
    \item InfoBench: MIT License
    \item AlpacaEval: Apache 2.0
    \item MT-Bench: Apache 2.0
    \item GPT-4 outputs: since we use this for research, we are in compliance with the GPT-4 terms of service. 
    \item Claude outputs: since we use this for research, we are in compliance with the Claude terms of service.
\end{itemize}

\end{document}